\documentclass[conference]{IEEEtran}
                                                          
\usepackage{style}

\FGfinalcopy

\usepackage{graphics} 
\usepackage{epsfig} 
\usepackage{mathptmx} 
\usepackage{times} 
\usepackage{amsmath} 
\usepackage{amssymb}  
\usepackage{etoolbox}

\makeatletter
\newcommand{\linebreakand}{%
  \end{@IEEEauthorhalign}
  \hfill\mbox{}\par
  \mbox{}\hfill\begin{@IEEEauthorhalign}
}
\makeatother

\begin{document}
\pagestyle{plain}

\title{\LARGE \bf
Unsupervised Compound Domain Adaptation for Face Anti-Spoofing 
}

\author{
\IEEEauthorblockN{Ankush Panwar}
\IEEEauthorblockA{University of Zurich \& ETH Zurich\\
{\tt\small ankush.panwar@uzh.ch}}
\and
\IEEEauthorblockN{Pratyush Singh}
\IEEEauthorblockA{University of Zurich \& ETH Zurich\\
{\tt\small psingh@student.ethz.ch}}
\and
\IEEEauthorblockN{Suman Saha}
\IEEEauthorblockA{ETH Zurich \\
{\tt\small suman.saha@vision.ee.ethz.ch}}
\linebreakand 
\IEEEauthorblockN{Danda Pani Paudel}
\IEEEauthorblockA{ETH Zurich \\
{\tt\small paudel@vision.ee.ethz.ch}}
\and
\IEEEauthorblockN{Luc Van Gool}
\IEEEauthorblockA{KU Leuven \& ETH Zurich \\
{\tt\small vangool@vision.ee.ethz.ch}}
}

\maketitle

\begin{abstract}

We address the problem of face anti-spoofing which aims to make the face verification systems robust in the real world settings. 
The context of detecting live vs. spoofed face images may differ significantly in the target domain, when compared to that of labeled source domain where the model is trained. Such difference may be caused due to new and unknown spoof types, illumination conditions, scene backgrounds, among many others. These varieties of differences make the target a compound domain, thus calling for the problem of the unsupervised compound domain adaptation. We demonstrate the effectiveness of the compound domain assumption for the task of face anti-spoofing, for the first time in this work.
To this end, we propose a memory augmentation method for adapting the source model to the target domain in a domain aware manner, inspired by~\cite{ocda}. The adaptation process is further improved by using the  curriculum learning and the domain agnostic source network training approaches. 
The proposed method successfully adapts to the compound target domain consisting multiple new spoof types. Our experiments on multiple benchmark datasets demonstrate the superiority of the proposed method over the state-of-the-art. Our source code will be made publicly available.

\end{abstract}

\section{INTRODUCTION}

The applications of facial recognition systems are ubiquitous and extensively used in our daily lives. Some examples include, the unlocking of smartphones, biometric payment or attendance systems. To facilitate the wide spread usage of face recognition systems, robust face verification is necessary. In this process, the liveliness of the presented face needs to be ensured prior to the recognition.Easy access to images of human faces makes the facial recognition systems vulnerable to spoof attacks such as a 3D mask, print, and video replay attacks~\cite{attacks}. The problem of face anti-spoofing ({FAS}) deals with detecting such attacks and is the key to preventing security breaches in biometric recognition applications. It has recently garnered increasing interest from the computer vision community~\cite{saha2020domain, shao2019, liu2018learning, tian2020face, ming2020survey}.

\begin{figure}
    \centering
    \includegraphics[width=\linewidth]{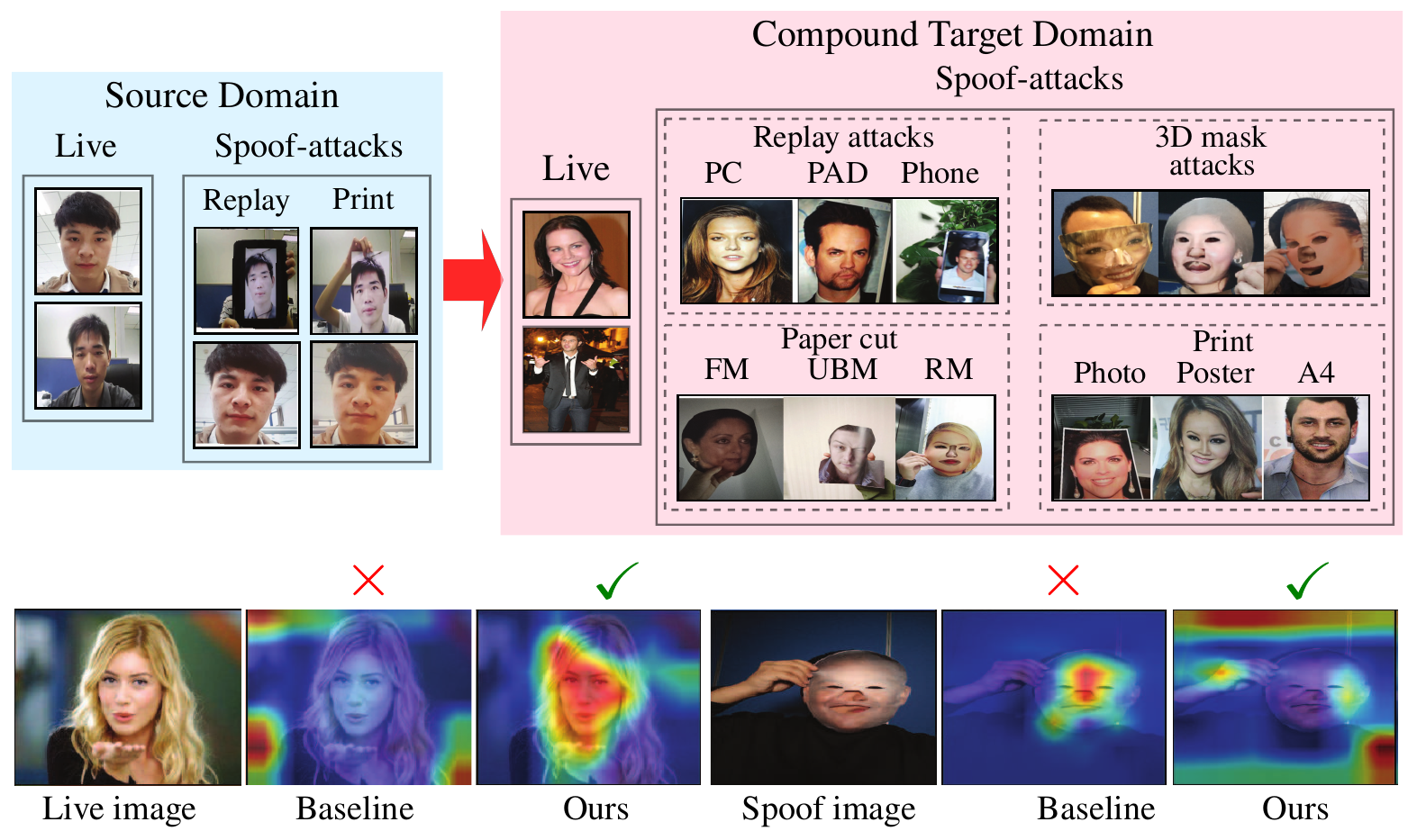}
    \caption{\textbf{Single labeled source and the unlabeled  compound target domains for FAS.} Top row -- unlike source domain, our compound target domain consists of multiple complex continuous sub-domains representing diverse spoof attacks, illuminations, backgrounds, capturing devices \textit{etc}. Such compound target domain assumption is better suited for the face anti-spoofing in the  real world settings. Bottom row -- visual comparison of the proposed model predictions (as Grad-CAM activation maps \cite{Selvaraju_2019}) with the baseline. The cross mark denotes an incorrect prediction and the check mark denotes a correct prediction. Note, the proposed model focuses on facial region for live samples, whereas for spoof, it focuses on different spoofing cues like hand and mask material. FM: face mask; UBM: upper body mask; RM: region mask.}
    \label{fig:teaser}
\end{figure}

Most existing FAS methods focus on supervised settings~\cite{Menotti_2015, yang2014learn, lilei}, where the test and the labeled training data are assumed to be from the same distribution \cite{Vapnik1998}. Unfortunately, such assumption does not hold true in practice -- due to the difference in spoof types, backgrounds, illumination conditions, among many others. Some methods address this problem by adapting the models trained on the labeled source data (or domain) to the target domain, with the help of unlabeled target data, by addressing the so-called the \emph{``Domain-shift"} problem~\cite{li2018b}. Other methods aim to train models that generalize across domains with the help of multi-domain labeled source data~\cite{saha2020domain, shao2019}. The domain adaptation methods assume the known target domain labels or treat the target as a single domain. On the other hand, domain generalization methods -- which remain agnostic about the target data domains -- requires the known domain labels in the source. In practice, neither the labeled multiple source nor the domain labeled target data may be available. Under such circumstances, one is left out with a singe domain labeled source  and  a composition of unknown multiple domains target data. Addressing the domain shift problem in such cases is known as the \emph{unsupervised compound domain adaptation}~\cite{ocda}. An illustration of such adaptation for FAS is shown in Fig.~\ref{fig:teaser}. 

This paper aims at answering the question of, \emph{how well does the compound target domain assumption hold for FAS?} In this context, various new unknown spoof types are assumed to exist in the target domain, in addition to the domain shift caused due to other factors such as illumination and background. To solve the problem of domain shift in compound target domain for FAS, we transfer the knowledge from source to the target domain dynamically in a step-by-step learning manner. Furthermore, to improve the adaptation on the unseen target domains, we employ a memory-based source-to-target knowledge transfer and also leverage the domain-specific attributes that are shared across the live and spoof samples from the source and target domains. The task is to learn an encoder network on the source dataset with known labels to distinguish between live and spoof images and later adapt the network to a more complex compound target domain with unknown labels.

The major contributions of this paper are summarised as:
\begin{itemize}
   \item We study the problem of unsupervised compound domain adaptation for face anti-spoofing, for the first time.  
   \item We propose a novel unsupervised domain adaptation framework for face anti-spoofing, inspired by~\cite{ocda}, tailored to the real-world FAS problems. 
    \item The proposed method achieves the state-of-the-art results on the challenging CelebA-Spoof~\cite{CelebA-Spoof} benchmark, in the context of unsupervised domain adaptation.
\end{itemize}

\section{Related Works}

\textbf{Hand-crafted-based traditional approaches.} The traditional FAS methods can be largely divided into two methods: \textit{texture-based methods} and \textit{temporal-based methods}. These methods leverage the hand-tuned features to differentiate the differences in texture between a real/spoof image. The most common approaches include HoG \cite{hog}, LBP \cite{LBP}, DoG \cite{DoG}, SIFT \cite{SIFT}, SURF \cite{surf}. The temporal-based methods leverage the ``liveness" of the image to differentiate the real image from the fake image~\cite{lip-motion, eye-blink}. The features generated with these methods are trained with traditional classifiers like SVM for classification. Other methods include transformation to the temporal domain \cite{opticalflow} and different color spaces \cite{colorspace}. These methods fail to generalize to different datasets since the learned ``texture" cues are very specific to a particular dataset and these learned features differ a lot in real-world settings.

\textbf{Deep learning based approaches.} CNN based deep learning methods for FAS have shown significant increase in the performance compared to the hand-crafted-based traditional methods~\cite{yang2014learn,jourabloo2018face_CNN,atoum}. The temporal specific features like Haralick features \cite{haralick} and optical flow \cite{opticalflow} have been proposed using CNNs. Feng \textit{et al.} \cite{feng} combines image quality information and motion information from optical flow with neural network for classification. In \cite{xu}, Xu \textit{et al.} propose an LSTM-CNN architecture to utilize temporal feature information for binary classification. Some researchers, Liu \textit{et al.} 2016 \cite{liuppg} and Liu, Jourabloo, and Liu 2018 \cite{liu2018learning} make use of the discriminative rPPG signals as temporal features.

The above-mentioned works perform well for intra-dataset testing but fail to generalize to unseen domains i.e. in the case of cross-dataset testing. This is mainly because of the extracted domain-biased features. Hence, the need to incorporate domain adaptation techniques become of great importance.

\textbf{Domain adaptation (DA) and Domain generalization (DG) based approaches.} To tackle the problem of domain shift between training and testing domains, DA \cite{da1}, \cite{da2}, \cite{ganin2016domainadversarial} and DG \cite{dg1}, \cite{dg2}, \cite{dg3}, \cite{dg4} techniques are widely used to learn domain-invariant features. Both methods attempt to bridge the gap between source and target domain with the difference that domain adaptation can leverage unlabeled target information. Li \textit{et al.} \cite{li2018b} and Wang \textit{et al.} \cite{wang} make the learned feature space domain invariant by minimizing the Maximum Mean Discrepancy (MMD) \cite{mmd} and adversarial training \cite{goodfellow2014generative} respectively. MMD computes the norm of the difference between two domain means. Yang \textit{et al.} \cite{yang11} suggested a subject-dependent transformation approach for synthesizing fake face features based on the premise that the relationship between real and fake samples belonging to a single subject can be expressed as a linear transformation. In practice, however, the major factors in facial image capturing like capturing medium, lighting conditions, angle type can be very different. This also motivates us to introduce unsupervised domain adaptation approach for FAS. From the perspective of domain generalization, Shao \textit{et al.} \cite{shao2019}, focuses on improving the generalization potential of FAS methods. Multiple feature extractors were taught to learn a generic feature space using adversarial learning. They also included an additional face depth supervision to improve the generalization potential even further. Jia \textit{et al.} \cite{jia2020singleside} propose a single-side domain generalization framework for FAS where the real faces from different domains are indistinguishable but the same is not applicable for spoof faces. Saha \textit{et al.} \cite{saha2020domain} learn the spatio-temporal features in an domain agnostic manner by using the gradient reversal layer introduced in \cite{ganin2015GRL} and class conditional domain discriminator module on the image-based and video-based network.

\section{Method} \label{MethodSection}

\begin{figure*}
    \includegraphics[width=\textwidth]{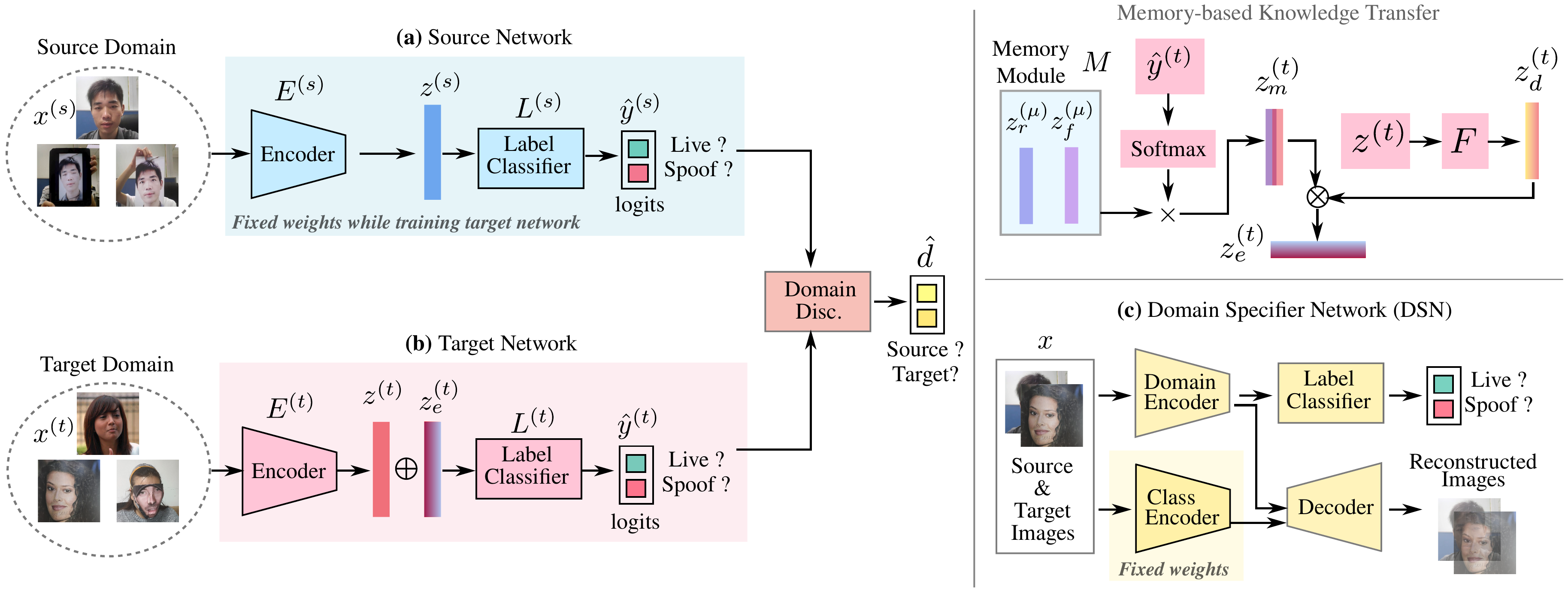}
    \caption{Overview of the different components of our proposed approach: (a,b) represent our target network together with memory module (top-right corner) and (c) provides the Domain Specifier Network (DSN) details. See \S~\ref{MethodSection} for more detailed explanation.}
    \label{fig:overview}
\end{figure*}

\subsection{Overview} \label{subsec:overview}
Fig. \ref{fig:overview} shows an overview of the the proposed deep neural network for FAS which comprises of the following blocks -- source network \textbf{(a)}, target network \textbf{(b)} and domain specifier network (DSN) \textbf{(c)}.
The network training is done in four stages.
\textbf{A:} First, we train the source network on source domain samples for label (live/spoof) classification task using a supervised cross-entropy loss (\S~\ref{subsec:sup_learning}).
\textbf{B:} Next, we initialize the target network with the pretrained-source model weights (obtained in Stage 1), and train it for domain alignment using a domain discriminator following an adversarial training (\S~\ref{subsec:adv_learning}). 
During the training of the target network using an adversarial objective, we enhance the learned representation by augmenting it with 
source domain's mean live and spoof features in a domain-aware manner.
For this, we employ a memory-based source-to-target knowledge transfer technique (\S~\ref{sec:memory_based_transfer}).
We show that our memory-based knowledge transfer helps to improve the FAS performance on multiple unseen spoof types (of target domain), e.g., mask attacks.
We do not update the model parameters of the source network during the target network training and keep them constant.
\textbf{C:} Furthermore, we train a DSN (\S~\ref{subsec:cl_for_da}) to learn domain-specific attributes from both source and target images.
The trained DSN is used to rank the source and target samples based on their $L_{2}$ distance in the domain-aware feature space.
The ranking is used to identify easy and hard examples for curriculum learning.
\textbf{D:} Finally, we again train the target network using curriculum learning (based on the DSN-based ranking) to improve the representation for FAS further.

\subsection{Notations} \label{prob-formulation}
Let $\mathfrak{D}^{(s)}$ and $\mathfrak{D}^{(t)}$ denote the source and the compound target domains;
$(x^{(s)}, y^{(s)})$ and $x^{(t)}$ represent samples from the source and target domains;
where $x \in \mathbb{R}^{H \times W \times 3}$ are color live and spoof face images,
$ y^{(s)}$ denotes the ground truth class labels (live or spoof) for source images.
The target domain comprises of multiple homogeneous domains $\mathfrak{D}^{(t)} = \{\mathfrak{D}^{(t)}_{1} \dots \mathfrak{D}^{(t)}_{U}\}$,
where the number of such homogeneous domains $U$ and the domain labels are unknown.
Furthermore, our proposed model consists of a source and target network each having a backbone encoder 
followed by a label classifier (Fig. \ref{fig:overview}). 
Let $E^{s}$ and $E^{t}$ are the encoders of the source and target networks respectively.
Similarly, $L^{s}$ and $L^{t}$ are the label classifiers.
The logits predicted by the label classifiers of the source and target networks are
denoted as $\hat{y}^{(s)}$, and $\hat{y}^{(t)}$ respectively, where $\hat{y} \in  \mathbb{R}^{2}$.
Besides, for adversarial learning, we use a domain discriminator which is denoted as $Disc$.
For simplicity, we remove the mini-batch dimension in all our notations.

\subsection{Supervised learning on source domain} \label{subsec:sup_learning}
In a classical domain adaptation setting, discriminative  features are learned by training the network on labeled source datasets. These features are then transformed to adapt to a different unlabeled dataset (target dataset) to boost the performance on this unseen target domain. In our first network, these class discriminative features are learned by training the source network
(Fig. \ref{fig:overview}(a))
in a supervised learning setup by optimizing the cross-entropy loss. Our source network is comprised of a CNN based encoder and a classifier with few fully connected neural network layers.

\subsection{Adversarial learning on target domain} \label{subsec:adv_learning}
We train the target network on both source and target domain samples using adversarial learning.
During the training optimization of the target network, we fuse the target representation with the source domain knowledge using a memory module which is explained in the following section.

\subsubsection{Memory-based Knowledge Transfer} \label{sec:memory_based_transfer}
We employ a memory-based source-to-target knowledge transfer \cite{ocda} to improve the adaptation ability of the learned FAS representation on unseen target domains (Fig. \ref{fig:overview} top-right).We perform this knowledge transfer by augmenting the target network's representation with the source features.
More specifically, we run inference using the trained source encoder $E^{(s)}$ (\S~\ref{subsec:sup_learning}) on the source domain samples and
extract all the live ${z}^{(s)}_{r} = E^{(s)}(x^{(s)}_{r})$ and spoof ${z}^{(s)}_{f} = E^{(s)}(x^{(s)}_{f})$ convolutional features from the last layer of the encoder network;
where ${z}^{(s)} \in \mathbb{R}^{D}$ is a $D$-dimensional feature vector.
We then compute the mean live ${z}^{(\mu)}_{r}$ and mean spoof ${z}^{(\mu)}_{f}$ features from these extracted features.
The mean features are jointly denoted as the memory module $M \in \mathbb{R}^{2 \times D}$.
\begin{equation}
\label{eqn:memory}
{z}^{(\mu)}_{c}=\frac{1}{N_c} \sum_{i_c=1}^{N_c} {z}^{(c)}_{i_c} , \quad
M = [{z}^{(\mu)}_{r}, {z}^{(\mu)}_{f}]
\end{equation}
where, $N_c$ is the number of images belongs to either live $r$ or spoof $f$ class, i.e., $c \in \{r,f\}$.
We use $M$ to augment the representation learned during training the target Encoder $E^{(t)}$ to transfer the knowledge from source to target domain.
The memory module helps the network learn how much information to retain from the target input and what extent of knowledge needs to be transferred from the source domain.
Such knowledge-transfer technique helps in better generalization on the unseen target domains for our compound domain adaptation scenario.
The knowledge is transferred from source to target domain in three steps which are presented in the following subsections. 

\textbf{Memory-augmented target feature.} The memory module $M$ in (\ref{eqn:memory}) contains class and domain-specific information from the source dataset,
and it is generated by computing the class centroids (i.e., the source live and spoof mean feature vectors).
Firstly, we transfer the knowledge of the source domain, which is stored in the memory module $M$ 
by fusing it with the predictions (logits) of the target network $\hat{y}^{(t)} = L^{(t)} ( E^{(t)} (x^{(t)}) )$ following (\ref{eq:enhancer}):
\begin{equation}
\label{eq:enhancer}
{z}^{(t)}_{m} = \text{Softmax}(\hat{y}^{(t)})^{T} M 
\end{equation}
The fusion is done by first converting the predicted logits $\hat{y}^{(t)}$ to softmax scores;
and then multiplying them with the memory module $M$.
Where, $\hat{y}^{(t)} \in \mathbb{R}^{1 \times 2}$ and $M \in \mathbb{R}^{2 \times D}$, 
and the resultant memory-augmented target feature ${z}^{(t)}_{m} \in \mathbb{R}^{1 \times D}$. 

\textbf{Domain-specific feature extraction.} Secondly, we train our target network to learn how much information to transfer from the source domain, i.e., from the memory-augmented target feature ${z}^{(t)}_{m}$;
and how much to retain from target domain, i.e., the representation learned by the target encoder $z^{(t)} = E^{(t)}(x^{(t)}) \in \mathbb{R}^{D}$.
To learn such domain-specific features, we employ a lightweight network $F$  as shown in (\ref{eq:domain_indicator})
\begin{equation}
\label{eq:domain_indicator}
z^{(t)}_{d} = F(z^{(t)}) 
\end{equation}
where, the resultant domain-specific feature $z^{(t)}_{d} \in \mathbb{R}^{D}$ is a $D$-dimensional feature vector. 

\textbf{Source-enhanced target feature.} Finally, we get the source-enhanced target feature $z^{(t)}_{e}$ of the target image $x^{(t)}$ 
which encodes information from both the source and target domain.
This is done by an element wise multiplication operation between the target domain-specifier $z^{(t)}_{d}$
and the  memory-augmented target feature ${z}^{(t)}_{m}$ as shown in (\ref{eq:source_enhanced_rep})
\begin{equation}
\label{eq:source_enhanced_rep}
z^{(t)}_{e} = z^{(t)}_{d}\otimes {z}^{(t)}_{m}
\end{equation}
It helps our network to dynamically calibrate how much knowledge to transfer from the source domain and
how much to rely on the target feature representation of the target image.
Intuitively, the larger domain gap between a target and the source domain, the more weight on the memory feature.  \\

\subsection{Curriculum learning for domain adaptation } \label{subsec:cl_for_da}
We further refine the learned representation of the target network by following a curriculum learning for incremental domain adaptation. 
The main idea of curriculum learning is first to identify the easy and hard training examples so that the target network can be trained in stages, i.e., starting with the easy examples and gradually feeding the network harder and harder examples as training progress.
We use the $L_{2}$ distance between the source and target domain-specific features to identify the easy and hard samples. 
The larger is the $L_{2}$ distance, the harder the samples are.
We employ a domain specifier network (DSN) to extract the domain-specific information from the source and target samples which is presented below.

\textbf{Domain Specifier Network.}
We use a DSN to leverage the domain-specific attributes shared across the live and spoof samples from the source and target domains.
The DSN is shown in Fig. \ref{fig:overview}\textbf{(c)}.
It has a domain encoder $E^{(d)}$, class (live or spoof) encoder $E^{(c)}$, label classifier $L^{(d)}$ and a decoder $G^{(d)}$.
The class encoder is initialized with the pretrained target encoder $E^{(t)}$ (\S~\ref{sec:memory_based_transfer}) 
weights and during DSN training its parameters are fixed. The pretrained target encoder captures primarily the class-discriminative representation,
which is domain agnostic. In other words, the features of the class encoder $E^{(c)}$ encodes class-specific information. 
On the other hand, we want the domain encoder to capture the domain-specific characteristics.
The assumption here is that the information which is not encoded by the class encoder preserves the domain-specific cues,
and the output of these two encoders ($E^{(d)}$ \& $E^{(c)}$) would provide sufficient information to reconstruct the input image.
Thus, we pass the output features of $E^{(d)}$ \& $E^{(c)}$ as inputs to the decoder $G^{(d)}$ 
which is trained to reconstruct the input image $x$ as $\hat{x}$, i.e., $\hat{x} = G^{(d)} ( E^{(d)} (x), E^{(c)} (x) ) \approx x $.

Training images from source and compound target datasets are passed as inputs to the DSN. 
The output of the domain encoder is fed as inputs to the label classifier and decoder, 
whereas the output of the class encoder is fed as input to the decoder only.
The label classifier predicts the class labels (live or spoof), 
and the decoder outputs a reconstructed version of the input image. 
The label classifier is trained with a cross-entropy loss with ground truth supervision on the source image.
For the target images, pseudo labels $y^{(t)}_{p}$ are used as ground truth supervision which are generated using the pretrained source network. 
We train the domain encoder and decoder of DSN using class-confusion and image reconstruction loss as in \cite{ocda}.

\textbf{Curriculum learning.} Finally, for curriculum learning, We use domain features extracted from DSN to rank our target samples.
We find domain distance between target and source images following (\ref{eqn:domaindist}).
\begin{equation}
\label{eqn:domaindist}
D_i=mean_m(\lVert E^{(d)}(x_i^{(t)}) - E^{{(d)}}(x_m^{(s)}) \rVert_2)
\end{equation}
We train the network by gradually recruiting more instances at every epoch that are far from source domain in feature space. This ensures a progressive feature learning process from simpler to complex images.

\subsection{Optimization Objectives} \label{subsec:objectives}
\textbf{Supervised learning on the source domain.} 
We train the source network for label classification task on the source domain's live and spoof training images $x^{(s)}$
using the ground truth labels $y^{(s)}$.
The parameters of the source network $\theta_{E^{(s)}}$ and $\theta_{L^{(s)}}$ (parameterizing $E^{(s)}$ and $L^{(s)}$),
collectively denoted as $\theta_{net^{(s)}}$, are learned to minimize the following supervised objective on the source domain.
\begin{equation}
\label{eq:sup_label_cls_loss_target}
\underset{\theta_{net^{(s)}}}{\text{min}} \quad - \sum_{i} y^{(s)}_{i} \log(\hat{y}^{(s)}_{i})
\end{equation}
\textbf{Adversarial learning for domain alignment.}
Our model relies on adversarial learning to align the feature distribution of the source and target domains. 
The parameters of the target network $\theta_{E^{(t)}}$ and $\theta_{L^{(t)}}$ (parameterizing $E^{(t)}$ and $L^{(t)}$), are collectively denoted as $\theta_{net^{(t)}}$.
For aligning the source and target feature distributions, we train the proposed target network and the domain discriminator $Disc$ (parameterized by $\theta_{Disc}$) following an adversarial training strategy. More precisely, the discriminator is tasked to correctly identify the sample domain being either source or target given the predicted logits $\hat{y}^{(s)}$ and $\hat{y}^{(t)}$ from the source and target networks following the objective:
\begin{equation}
\label{eq:adv_loss_disc1}
\underset{\theta_{Disc}}{\text{min}} \quad - \sum_{i} d_{i} \log(\hat{d}_{i})
\end{equation}
At the same time, the target network parameters are learned to maximize the domain classification loss by fooling the discriminator on the target samples using the following adversarial objective:
\begin{equation}
\label{eq:adv_loss_disc2}
\underset{\theta_{net^{(t)}}}{\text{min}} \quad - \sum_{i} d^{f}_{i} \log(\hat{d}_{i})
\end{equation}
where, subscript $i$ is the sample index, $d$ and $\hat{d}$ are the ground truth and predicted domain labels (source or target),
$d^{f}$ is the fake domain label ``source'' assigned for target samples to confuse the discriminator.

\section{Experiments and results}

\subsection{Datasets}

\begin{table*}[t]
    \begin{center}
    \caption{Significant differences between source (Ca, Ms \& Ou) and  Target (Ce) dataset in terms of spoof types, amount of data, subjects and sensors }
    \label{tab:Dataset_comparison}
     \begin{tabular}{|c|c|c|c|c|c|} 
     \hline
     Source Datasets & Subjects & Data & Sensors & Spoof Types  & Real/Spoof\\
     \hline
     \textbf{CASIA-MFSD (Ca)} & 50 & 600 videos & 3 & 1 Print, 1 Video-replay & 1:3\\ 
     \hline
     \textbf{MSU-MFSD (Ms)} & 35 & 440 videos & 2 & 1 Print, 2 Video-replay & 1:3\\
     \hline
     \textbf{OULU-NPU (Ou)} & 55 & 5,940 videos & 6 & 1 Print, 2 Video-replay & 2:3\\
     \hline
     \textbf{CelebA-Spoof (Ce)} & 10,177 & 625,537 images & $>$10 & 3 Print, 3 Replay
    1 3D, 3 Paper Cut & 1:3\\
     \hline
    \end{tabular}
    
    \end{center}
\end{table*}

For experiments, we use widely known FAS benchmark datasets Oulu-NPU \cite{oulu} (Ou for short), CASIA-MFSD \cite{CASIA} (Ca for short), MSU-MFSD \cite{MSU} (Ms for short) as our source data. However, these FAS datasets are limited in both quantity and diversity as listed in Table \ref{tab:Dataset_comparison}. Ms, Ca, Ou suffer from lacking sufficient spoof types, subjects, sessions and input sensors. Our compound target dataset CelebA-Spoof \cite{CelebA-Spoof} (Ce for short) is a large scale FAS dataset with rich and diverse annotations (see Table \ref{tab:Dataset_comparison}). It comprises of 625,537 pictures of 10,177 subjects which is significantly larger than our combined source datasets as shown in Table \ref{tab:Dataset_comparison}. 
Ce has 4 different spoof types (print, paper-cut, replay and 3D mask) captured in different lighting conditions, environment, device, angle capture, material \textit{etc.} Few of these realistic spoof attacks in Ce such as paper-cut and 3D mask attack is not present in the MsCaOu source datasets. Ce is also the first FAS dataset covering spoof images in the outdoor environment whereas source datasets only contains indoor image. The enormous diversity in the compound target dataset, compared to the source datasets, creates a greater challenge to our task of unsupervised compound domain adaptation for FAS. 

\subsection{Evaluation metrics} 
Half Total Error Rate (HTER) is the most common evaluation metric for face anti-spoofing task. Therefore, we report the performance of our network using HTER \cite{hter}. We also provide the ROC curve and Area Under Curve (AUC) as our evaluation metrics for further assessment.

\subsection{Implementation details} \label{implementation}
We use ResNet-50 \cite{resnet50} as our network encoder in source network. The dimension of the input images is (224×224). The source network is trained for 10 epochs with a learning rate of 0.0001 and weight decay of 0.00001 using ADAM optimizer. The size of the encoder output feature dimension is 2048. The target network with memory augmentation is a copy of the source network with weight initialized from the previously trained source network. It is also trained with the same weight decay but a slower learning rate of 0.000001 of ADAM optimizer. For the target network, an alternating optimization routine is used to optimize the discriminator with domain classification cross entropy loss and the target network with domain confusion loss (refer  \S~\ref{subsec:objectives}). During the training of our Domain Specifier Network (DSN), we used the ADAM optimizer with a learning rate of 0.00001 and used class-confusion loss and reconstruction loss (L1 loss) to optimize the network. In the final stage, we further train our target network with curriculum learning and we use a learning rate of 0.000001 and weight decay of 0.00001 with a total of 10 epochs. The mini-batch size is 16 for all the networks while training and 32 while testing on the target domain.

We carried out two sets of experiments and details of these experiments are listed below:
\begin{enumerate}
  \item Our \textbf{SSD (Single Source Domain)} Training: In this experiment, our assumption is that the combined Ms, Ca and Ou datasets represent one source domain. With this assumption, our source network (\S~\ref{subsec:sup_learning}) was trained in a simple supervised setting using cross entropy loss where class labels (live/spoof) of the source dataset are known. The whole experiment was conducted in a vanilla setting where every network training was performed as explained in \S~\ref{MethodSection}. 
  \item Our \textbf{MSD (Multiple Source Domain)} Training: In this experiment, we considered each source dataset as one domain, and therefore our source dataset contains 3 source domains (Ms, Ca \& Ou). To extract rich classification features, we trained our source network in a domain-agnostic manner \cite{saha2020domain} to get domain independent classification features from source datasets by utilizing source domain labels. However, the rest of our training process after Source Network is exactly the same as explained in \S~\ref{MethodSection}.
\end{enumerate}

\begin{table}
\begin{center}

\caption{\textbf{Performance in SSD training setting}: Comparison of our proposed model with oracle and common face anti-spoofing domain adaptation method on CelebA-spoof \cite{CelebA-Spoof} test set. Source datasets \textbf{MsCaOu} are considered as single domain in these experiment. See \S~\ref{results} for more details.}
\label{tab:experiment1}
\begin{tabular}{|c|c|c|}
\hline
 & \multicolumn{2}{|c|}{MsCaOu\textrightarrow  Ce} \\
\hline
\textbf{Methods} & \textbf{HTER (\%)} & \textbf{AUC (\%)}\\
\hline
Oracle(Supervised Learning) & 13.4 & 98.7\\
\hline
Source Network & 37.6 & 67.5\\
\hline
GRL Layer \cite{ganin2015GRL} & 29.1 & 76.4\\
\hline
Domain-confusion \cite{tzeng2017adversarial} & 33.7 & 70.3\\
\hline
\textbf{Our (SSD Training)} & \textbf{26.1} & \textbf{80.0}\\
\hline
\end{tabular}
\end{center}

\end{table}

\begin{table}
\begin{center}
\caption{\textbf{Performance in MSD training setting}: Comparison of our proposed model to state-of-the-art face anti-spoofing generalization method on CelebA-spoof \cite{CelebA-Spoof} test set. Source datasets \textbf{MsCaOu} are considered as 3 distinct known domains in these experiments. See \S~\ref{results} for more details.}
\label{tab:experiments2}
\begin{tabular}{|c|c|c|}
\hline
 & \multicolumn{2}{|c|}{MsCaOu\textrightarrow  Ce} \\
\hline
\textbf{Methods} & \textbf{HTER (\%)} & \textbf{AUC (\%)}\\
\hline
Saha \textit{et al.} \cite{saha2020domain} & 27.1 & 79.2\\
\hline
\textbf{Our (MSD Training)} & \textbf{20.06} & \textbf{85.8}\\
\hline
\end{tabular}
\end{center}
\end{table}

\begin{figure}
    \centering
    \includegraphics[width=8 cm]{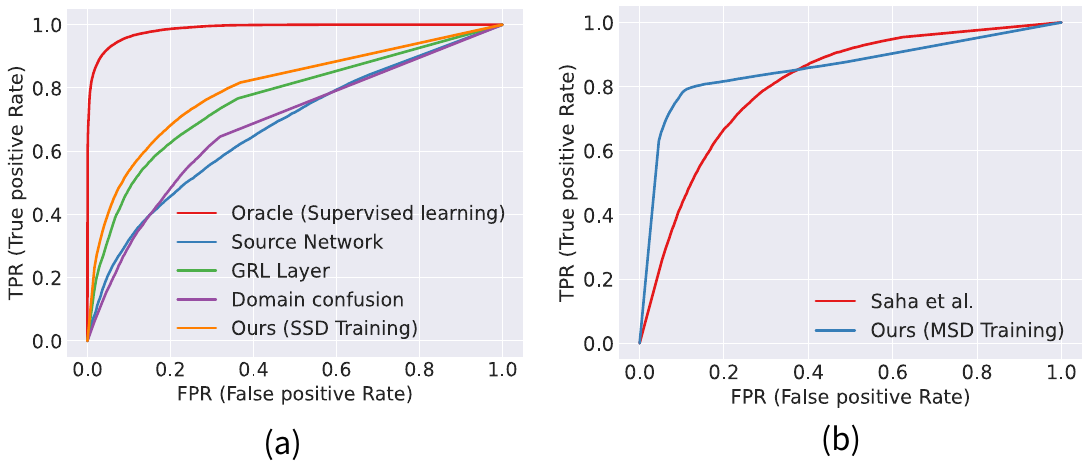}
    \caption{ROC (Receiver operating characteristic) curve for comparison of common domain adaptation and domain generalization methods with our proposed architecture with SSD training and MSD training respectively in (a) \& (b). }
    \label{fig:roccurves}
\end{figure}

\begin{table*}
\begin{center}
\caption{Prediction Error for each spoof type in target test data (CelebA-Spoof) from our proposed architecture with SSD and MSD training.}
\label{tab:spooferror1}
\begin{tabular}{|c|c|c|c|c|c|c|c|c|c|c|c|c|}
\hline
\multicolumn{2}{|c|}{\textbf{Spoof Type}} & \multicolumn{3}{|c|}{\textbf{Print}} & \multicolumn{3}{|c|}{\textbf{Paper-cut}} & \multicolumn{3}{|c|}{\textbf{Replay}} & \textbf{3D-Mask}\\
\hline
\multicolumn{2}{|c|}{} & Photo & {Poster} & {A4} 
& {Face Mask}
& {Upper Body Mask}
& {Region Mask}
& {PC}
& {Pad}
& {Phone}
& {3D Mask} \\
\hline
\multicolumn{2}{|c|}{\textbf{Test Images}} & 3600	& 5420 & 6083 & 4287 & 6097 & 3530 & 6477 & 3659 & 4483 & 3610\\
\hline
\multirow{2}{*}{\textbf{\shortstack{Prediction \\ Error \%}}} & \textbf{ Our SSD } & 18.14 & 21.55 & 33.13 & 21.53 & 18.12 & 16.70 & 45.73 & 45.48 & 28.55 & 16.73\\
\cline{2-11}
& \textbf{ Our MSD} & 21.97 & 9.61 & 20.19 & 29.69 & 19.32 & 10.40 & 38.89 & 21.75 & 24.89 & 19.64\\
\hline

\end{tabular}
\end{center}
\end{table*}

\subsection{Result comparison} \label{results}

To the best of our knowledge, we couldn't find any work which performed the experiments by taking CelebA-Spoof dataset as the evaluation set in domain adaptation or domain generalization setting. For this sole reason, we provide our own baseline models by performing experiments on common domain adaptation methods to compare the results with our SSD trained model (Table \ref{tab:experiment1} \& Fig. \ref{fig:roccurves}(a)).
\begin{enumerate}
    \item Oracle: In this experiment, the ResNet model is trained by performing supervised learning on our target dataset. Ideally, it should give us the best performance on our target dataset since target labels were utilized which is not used in our proposed architecture.
    \item Source Network: We evaluated our target dataset on our source network which is trained in supervised learning setting on source datasets.
    \item GRL Layer: It follows the Ganin \textit{et al.} \cite{ganin2015GRL} architecture where gradient reversal layer (GRL) was used to adapt the network to the target domain.
    \item Domain Confusion: The architecture consists of domain confusion adversarial loss as in Tzeng \textit{et al.} \cite{tzeng2017adversarial} to train the model in a domain adaptive manner.
\end{enumerate}

We also compared our MSD trained model with the current the state-of-the-art in FAS domain shift problem by Saha \textit{et al.} \cite{saha2020domain}. It learns domain agnostic classification features using a gradient reversal layer (GRL) and a class conditional domain discriminator. 

 Our SSD experiment performed significantly better than all the baseline architectures and very competitive compared to our oracle as shown in Table \ref{tab:experiment1} and Fig. \ref{fig:roccurves}(a). Moreover, our MSD experiment outperformed the state-of-the-art Saha \textit{et al.} \cite{saha2020domain}) on CelebA-Spoof compound target dataset (Table \ref{tab:experiments2} \& Fig. \ref{fig:roccurves}(b)). The significantly better performance mostly lies in our network's ability to dynamically transfer knowledge from source to target domain (see Fig. \ref{fig:tsne}). It utilizes memory module to mitigate the compound target domain problem in face anti-spoofing task. We also reported prediction errors of each spoof type in Table \ref{tab:spooferror1} for our SSD and MSD experiments. It can be seen that our network is able to classify the print, paper-cut and 3D mask attacks with high accuracy. However, our performance degraded for video replay attack (especially on PC).

\subsection{Visualization of CNN learned features}

\begin{figure*}
    \centering
    \includegraphics[width=15cm]{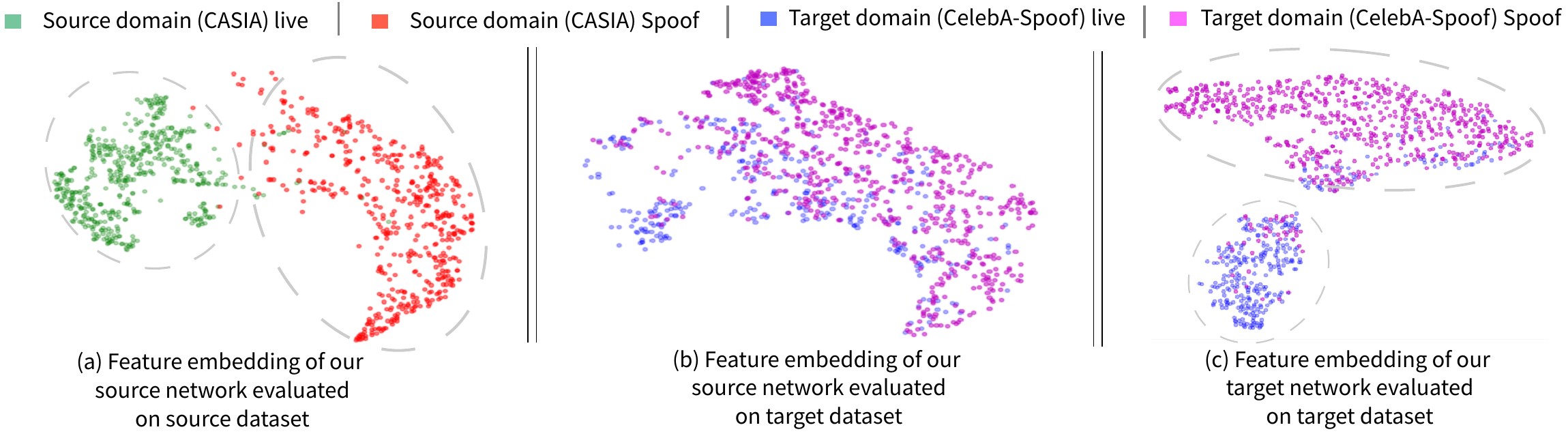}
    \caption{The t-SNE plots of CNN features from last convolution layer of our Source encoder (a,b) vs our target encoder (c). It shows how source network in (b) is ineffective in distinguishing spoof and live images in our target dataset. While our adapted target network (c) created clear live and spoof clusters in feature space.}
    \label{fig:tsne}
\end{figure*}

For more detailed evaluation, the variables embedded in the highest layer of our network encoder were visually evaluated after two-dimensional reduction using the t-distributed stochastic neighbour embedding (t-SNE) technique in Fig. \ref{fig:tsne}. Fig. \ref{fig:tsne}(a) depicts that our supervied source network is able to distinguish between live and spoof images of the source dataset (only CASIA is taken for simplicity) in feature space. However our source network fails to classify our unseen unlabeled compound target dataset (CelebA-spoof) in Fig. \ref{fig:tsne}(b). Fig. \ref{fig:tsne}(c) shows our proposed target network evaluated on compound domain CelebA-spoof dataset. It can be seen in Fig. \ref{fig:tsne}(c) that live and spoof features of the target domains are far apart in feature space. Our target network interestingly adapts and learns better representation of the compound target domain’s live and spoof features. From these visualizations, we can conclude that our network adapts to our compound target dataset irrespective of the large domain gap.

\subsection{Class activation map visualization}
\begin{figure}
    \centering
    \includegraphics[width=8 cm]{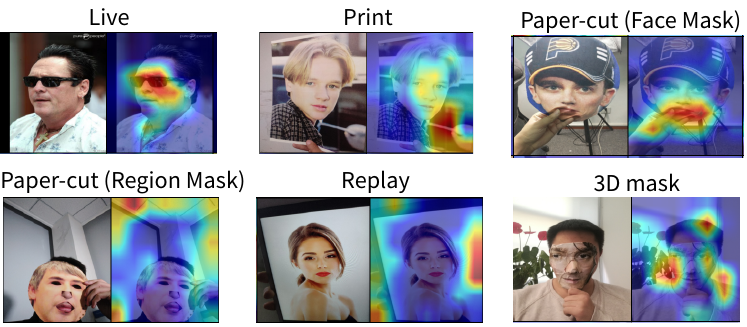}
    \caption{Activation map visualization of the proposed target network on CelebA-spoof \cite{CelebA-Spoof} dataset samples. For each live/spoof type, we have shown the original image and its associated network class activation maps using Grad-CAM \cite{Selvaraju_2019}}
    \label{fig:gradcam}
\end{figure}

To study and visualize the class activation maps to get a better understanding of features learned by network while making a particular prediction decision for live and spoof examples. We used the Gradient-weighted Class Activation Mapping (Grad-CAM) \cite{Selvaraju_2019} technique to produce a coarse localization map highlighting important regions in the image for class prediction. In Fig. \ref{fig:gradcam}, we show the class activation map for live, print, paper-cut (face mask and region mask), video-replay and 3D mask attack test samples. In Fig. \ref{fig:gradcam}, network focuses on facial region for prediction of live samples and is intuitive since most of the live image information comes from face of the person such as eyes, mouth, nose, skin \textit{etc.}. However while predicting label for spoof attacks,  it gives importance to print paper, video screen, edges of the device, glare on the device, background objects, hands holding the mask/paper-cut and edges of the mask as shown in Fig. \ref{fig:gradcam}.

\subsection{Ablation Study}
\begin{figure}
    \centering
    \includegraphics[width=8cm]{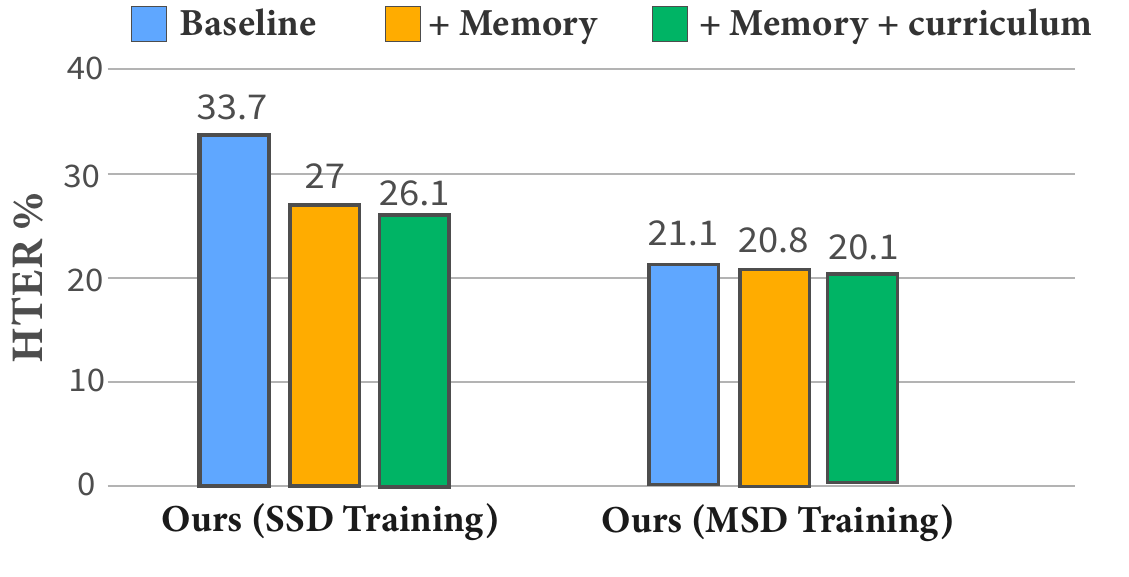}
    \caption{Ablation study of our proposed target network. Results are shown in HTER \%. It shows the performance contribution of memory augmentation and curriculum learning in our target network with SSD and MSD training experiments.}
    \label{fig:ablation}
\end{figure}

In this section, we provide more detailed experimental results to validate the efficacy of the obtained results.

As shown in Fig. \ref{fig:ablation} for both of our experiments, the target network trained without any memory module (baseline) performs the worst. The major performance boost comes from the memory module in our SSD experiment which provided 6.7\% improvement in HTER. Moreover, incorporating curriculum learning approach improved our HTER by 0.9\%. However in MSD Experiment, memory module did not provide significant improvement over the baseline. This result was expected since memory augmentation relies on domain gap between source and target datasets for transferring knowledge and  we are providing domain independent classification features for adaptation. Such features contains minimal information about domain in feature space and thus reducing the contribution of our memory module. Further, the curriculum training schedule in MSD experiment learned better classification features and boosted the performance by 0.7\%.

\section{CONCLUSION}

We addressed a real-world problem of domain shift in labeled single source and unlabeled compound target domains, in the context of face anti-spoofing. Diversity in factors such as spoof types, backgrounds, lighting conditions,
camera resolutions, capturing device, spoof materials makes CNN learned feature representation domain dependent. This leads to poor prediction performance on unseen domains and spoof types. We addressed these issues by proposing a domain adaptation method where we dynamically transfer knowledge from source to target domain in order to bridge the domain gap. We further improved our prediction performance by incorporating curriculum learning. We tested our network on a complex compound CelebA-spoof dataset, which was also used as an unlabeled target for the first time. We demonstrated state-of-the-art performance on CelebA-spoof dataset and also illustrated the qualitative improvement of the learned features using Grad-Cam and t-SNE plots.

{\small
\bibliographystyle{ieee}
\bibliography{bibliograph}

\begin{thebibliography}{10}\itemsep=-1pt

\bibitem{haralick}
A.~Agarwal, R.~Singh, and M.~Vatsa.
\newblock Face anti-spoofing using haralick features.
\newblock pages 1--6, 09 2016.

\bibitem{atoum}
Y.~Atoum, Y.~Liu, A.~Jourabloo, and X.~Liu.
\newblock Face anti-spoofing using patch and depth-based cnns.
\newblock 10 2017.

\bibitem{opticalflow}
W.~Bao, H.~Li, N.~Li, and W.~Jiang.
\newblock A liveness detection method for face recognition based on optical
  flow field.
\newblock {\em 2009 International Conference on Image Analysis and Signal
  Processing}, pages 233--236, 2009.

\bibitem{hter}
S.~Bengio and J.~Mari{\'{e}}thoz.
\newblock A statistical significance test for person authentication.
\newblock In {\em Proceedings of Odyssey 2004: The Speaker and Language
  Recognition Workshop}, number Idiap-RR-83-2003, 2004.

\bibitem{colorspace}
Z.~Boulkenafet, J.~Komulainen, and A.~Hadid.
\newblock face anti-spoofing based on color texture analysis, 2015.

\bibitem{surf}
Z.~{Boulkenafet}, J.~{Komulainen}, and A.~{Hadid}.
\newblock Face antispoofing using speeded-up robust features and fisher vector
  encoding.
\newblock {\em IEEE Signal Processing Letters}, 24(2):141--145, 2017.

\bibitem{oulu}
Z.~{Boulkenafet}, J.~{Komulainen}, L.~{Li}, X.~{Feng}, and A.~{Hadid}.
\newblock Oulu-npu: A mobile face presentation attack database with real-world
  variations.
\newblock In {\em 2017 12th IEEE International Conference on Automatic Face
  Gesture Recognition (FG 2017)}, pages 612--618, 2017.

\bibitem{feng}
L.~Feng, L.~Po, Y.~Li, X.~Xu, F.~Yuan, T.~C.-H. Cheung, and K.-W. Cheung.
\newblock Integration of image quality and motion cues for face anti-spoofing:
  A neural network approach.
\newblock {\em Journal of Visual Communication and Image Representation}, 38,
  04 2016.

\bibitem{da2}
B.~Fernando, A.~Habrard, M.~Sebban, and T.~Tuytelaars.
\newblock Unsupervised visual domain adaptation using subspace alignment.
\newblock 12 2013.

\bibitem{ganin2015GRL}
Y.~Ganin and V.~Lempitsky.
\newblock Unsupervised domain adaptation by backpropagation, 2015.

\bibitem{ganin2016domainadversarial}
Y.~Ganin, E.~Ustinova, H.~Ajakan, P.~Germain, H.~Larochelle, F.~Laviolette,
  M.~Marchand, and V.~Lempitsky.
\newblock Domain-adversarial training of neural networks, 2016.

\bibitem{dg2}
M.~Ghifary, D.~Balduzzi, W.~B. Kleijn, and M.~Zhang.
\newblock Scatter component analysis: {A} unified framework for domain
  adaptation and domain generalization.
\newblock {\em {IEEE} Trans. Pattern Anal. Mach. Intell.}, 39(7):1414--1430,
  2017.

\bibitem{goodfellow2014generative}
I.~J. Goodfellow, J.~Pouget-Abadie, M.~Mirza, B.~Xu, D.~Warde-Farley, S.~Ozair,
  A.~Courville, and Y.~Bengio.
\newblock Generative adversarial networks, 2014.

\bibitem{da1}
R.~{Gopalan}, {Ruonan Li}, and R.~{Chellappa}.
\newblock Domain adaptation for object recognition: An unsupervised approach.
\newblock In {\em 2011 International Conference on Computer Vision}, pages
  999--1006, 2011.

\bibitem{mmd}
A.~Gretton, A.~Smola, J.~Huang, M.~Schmittfull, K.~Borgwardt, and
  B.~Sch{\"o}lkopf.
\newblock {\em Covariate shift and local learning by distribution matching},
  pages 131--160.
\newblock MIT Press, Cambridge, MA, USA, 2009.

\bibitem{resnet50}
K.~He, X.~Zhang, S.~Ren, and J.~Sun.
\newblock Deep residual learning for image recognition, 2015.

\bibitem{jia2020singleside}
Y.~Jia, J.~Zhang, S.~Shan, and X.~Chen.
\newblock Single-side domain generalization for face anti-spoofing, 2020.

\bibitem{jourabloo2018face_CNN}
A.~Jourabloo, Y.~Liu, and X.~Liu.
\newblock Face de-spoofing: Anti-spoofing via noise modeling, 2018.

\bibitem{dg1}
A.~Khosla, T.~Zhou, T.~Malisiewicz, A.~A. Efros, and A.~Torralba.
\newblock Undoing the damage of dataset bias.
\newblock In {\em Proceedings of the 12th European Conference on Computer
  Vision - Volume Part I}, ECCV'12, page 158–171, Berlin, Heidelberg, 2012.
  Springer-Verlag.

\bibitem{lip-motion}
K.~{Kollreider}, H.~{Fronthaler}, M.~I. {Faraj}, and J.~{Bigun}.
\newblock Real-time face detection and motion analysis with application in
  “liveness” assessment.
\newblock {\em IEEE Transactions on Information Forensics and Security},
  2(3):548--558, 2007.

\bibitem{hog}
J.~Komulainen, A.~Hadid, and M.~Pietikainen.
\newblock Context based face anti-spoofing.
\newblock pages 1--8, 09 2013.

\bibitem{LBP}
J.~Komulainen, A.~Hadid, and M.~Pietikäinen.
\newblock Face spoofing detection from single images using micro-texture
  analysis.
\newblock 10 2011.

\bibitem{attacks}
S.~Kumar, S.~Singh, and J.~Kumar.
\newblock A comparative study on face spoofing attacks.
\newblock In {\em 2017 International Conference on Computing, Communication and
  Automation (ICCCA)}, pages 1104--1108, 2017.

\bibitem{dg3}
D.~Li, Y.~Yang, Y.-Z. Song, and T.~M. Hospedales.
\newblock Deeper, broader and artier domain generalization, 2017.

\bibitem{li2018b}
H.~{Li}, W.~{Li}, H.~{Cao}, S.~{Wang}, F.~{Huang}, and A.~C. {Kot}.
\newblock Unsupervised domain adaptation for face anti-spoofing.
\newblock {\em IEEE Transactions on Information Forensics and Security},
  13(7):1794--1809, 2018.

\bibitem{lilei}
L.~Li, X.~Feng, Z.~Boulkenafet, Z.~Xia, M.~Li, and A.~Hadid.
\newblock An original face anti-spoofing approach using partial convolutional
  neural network.
\newblock pages 1--6, 12 2016.

\bibitem{liuppg}
S.~Liu, P.~C. Yuen, S.~Zhang, and G.~Zhao.
\newblock 3d mask face anti-spoofing with remote photoplethysmography.
\newblock volume 9911, pages 85--100, 10 2016.

\bibitem{liu2018learning}
Y.~Liu, A.~Jourabloo, and X.~Liu.
\newblock Learning deep models for face anti-spoofing: Binary or auxiliary
  supervision, 2018.

\bibitem{ocda}
Z.~Liu, Z.~Miao, X.~Pan, X.~Zhan, D.~Lin, S.~X. Yu, and B.~Gong.
\newblock Open compound domain adaptation, 2020.

\bibitem{Menotti_2015}
D.~Menotti, G.~Chiachia, A.~Pinto, W.~Robson~Schwartz, H.~Pedrini,
  A.~Xavier~Falcao, and A.~Rocha.
\newblock Deep representations for iris, face, and fingerprint spoofing
  detection.
\newblock {\em IEEE Transactions on Information Forensics and Security},
  10(4):864–879, Apr 2015.

\bibitem{ming2020survey}
Z.~Ming, M.~Visani, M.~M. Luqman, and J.-C. Burie.
\newblock A survey on anti-spoofing methods for face recognition with rgb
  cameras of generic consumer devices, 2020.

\bibitem{dg4}
S.~Motiian, M.~Piccirilli, D.~A. Adjeroh, and G.~Doretto.
\newblock Unified deep supervised domain adaptation and generalization, 2017.

\bibitem{eye-blink}
G.~{Pan}, L.~{Sun}, Z.~{Wu}, and S.~{Lao}.
\newblock Eyeblink-based anti-spoofing in face recognition from a generic
  webcamera.
\newblock In {\em 2007 IEEE 11th International Conference on Computer Vision},
  pages 1--8, 2007.

\bibitem{SIFT}
K.~{Patel}, H.~{Han}, and A.~K. {Jain}.
\newblock Secure face unlock: Spoof detection on smartphones.
\newblock {\em IEEE Transactions on Information Forensics and Security},
  11(10):2268--2283, 2016.

\bibitem{saha2020domain}
S.~Saha, W.~Xu, M.~Kanakis, S.~Georgoulis, Y.~Chen, D.~P. Paudel, and L.~V.
  Gool.
\newblock Domain agnostic feature learning for image and video based face
  anti-spoofing, 2020.

\bibitem{Selvaraju_2019}
R.~R. Selvaraju, M.~Cogswell, A.~Das, R.~Vedantam, D.~Parikh, and D.~Batra.
\newblock Grad-cam: Visual explanations from deep networks via gradient-based
  localization.
\newblock {\em International Journal of Computer Vision}, 128(2):336–359, Oct
  2019.

\bibitem{shao2019}
R.~{Shao}, X.~{Lan}, J.~{Li}, and P.~C. {Yuen}.
\newblock Multi-adversarial discriminative deep domain generalization for face
  presentation attack detection.
\newblock In {\em 2019 IEEE/CVF Conference on Computer Vision and Pattern
  Recognition (CVPR)}, pages 10015--10023, 2019.

\bibitem{DoG}
X.~Tan, y.~Liu, J.~Liu, and L.~Jiang.
\newblock Face liveness detection from a single image with sparse low rank
  bilinear discriminative model.
\newblock volume 6316, pages 504--517, 09 2010.

\bibitem{tian2020face}
Y.~Tian, K.~Zhang, L.~Wang, and Z.~Sun.
\newblock Face anti-spoofing by learning polarization cues in a real-world
  scenario, 2020.

\bibitem{tzeng2017adversarial}
E.~Tzeng, J.~Hoffman, K.~Saenko, and T.~Darrell.
\newblock Adversarial discriminative domain adaptation, 2017.

\bibitem{Vapnik1998}
V.~N. Vapnik.
\newblock {\em Statistical Learning Theory}.
\newblock Wiley-Interscience, 1998.

\bibitem{wang}
G.~{Wang}, H.~{Han}, S.~{Shan}, and X.~{Chen}.
\newblock Improving cross-database face presentation attack detection via
  adversarial domain adaptation.
\newblock In {\em 2019 International Conference on Biometrics (ICB)}, pages
  1--8, 2019.

\bibitem{MSU}
D.~{Wen}, H.~{Han}, and A.~K. {Jain}.
\newblock Face spoof detection with image distortion analysis.
\newblock {\em IEEE Transactions on Information Forensics and Security},
  10(4):746--761, 2015.

\bibitem{xu}
Z.~{Xu}, S.~{Li}, and W.~{Deng}.
\newblock Learning temporal features using lstm-cnn architecture for face
  anti-spoofing.
\newblock In {\em 2015 3rd IAPR Asian Conference on Pattern Recognition
  (ACPR)}, pages 141--145, 2015.

\bibitem{yang2014learn}
J.~Yang, Z.~Lei, and S.~Z. Li.
\newblock Learn convolutional neural network for face anti-spoofing, 2014.

\bibitem{yang11}
J.~Yang, Z.~Lei, D.~Yi, and S.~Z. Li.
\newblock Person-specific face antispoofing with subject domain adaptation.
\newblock {\em IEEE Transactions on Information Forensics and Security},
  10(4):797--809, 2015.

\bibitem{CelebA-Spoof}
Y.~Zhang, Z.~Yin, Y.~Li, G.~Yin, J.~Yan, J.~Shao, and Z.~Liu.
\newblock Celeba-spoof: Large-scale face anti-spoofing dataset with rich
  annotations.
\newblock In {\em European Conference on Computer Vision (ECCV)}, 2020.

\bibitem{CASIA}
Z.~{Zhang}, J.~{Yan}, S.~{Liu}, Z.~{Lei}, D.~{Yi}, and S.~Z. {Li}.
\newblock A face antispoofing database with diverse attacks.
\newblock In {\em 2012 5th IAPR International Conference on Biometrics (ICB)},
  pages 26--31, 2012.

\end{thebibliography}
}

\end{document}